\documentclass{article}
\usepackage{ijcai18}

\usepackage{times}
\usepackage{xcolor}
\usepackage{soul}
\usepackage[utf8]{inputenc}
\usepackage[small]{caption}

\usepackage{graphicx}
\usepackage{hyperref}
\usepackage[vlined,ruled]{algorithm2e}
\usepackage{subcaption}
\usepackage{amsfonts,amsmath,amsthm, amssymb}

\usepackage[mathscr]{euscript}
\usepackage{multirow}
\usepackage{bm}

\title{Learning High-level Representations from Demonstrations}
\author{
Garrett Andersen ~~~~ 
Peter Vrancx ~~~~ 
Haitham Bou-Ammar 
\\ 
PROWLER.io \\
\{garrett,
peter,
haitham\} @prowler.io
}

\usepackage{bbm} 

\newcommand{\p}{p}  
\renewcommand{\S}{\mathscr{S}} 
\newcommand{\A}{\mathscr{A}} 
\renewcommand{\r}{r} 
\newcommand{\M}{\mathscr{M}} 

\newcommand{\V}{v}

\newcommand{\I}{\mathscr{I}} 

\newcommand{\R}{\mathcal{R}}

\renewcommand{\r}{r}

\newcommand{\E}{\mathbb{E}} 

\DeclareMathOperator*{\argmax}{arg\,max}

\newcommand{\1}{\mathbb{I} } 
\newcommand{\indic}[1]{\1\{#1\}} 

\newcommand{\Real}{\mathbb{R}}

\newcommand{\beq}{\begin{equation}}
\newcommand{\eeq}{\end{equation}}

\newcommand{\beqa}{\begin{eqnarray}}
\newcommand{\eeqa}{\end{eqnarray}}

\newcommand{\beqan}{\begin{eqnarray*}}
\newcommand{\eeqan}{\end{eqnarray*}}

\begin{document}

\maketitle

\begin{abstract}
Hierarchical learning (HL) is key to solving complex sequential decision problems with long horizons and sparse rewards. It allows learning agents to break-up large problems into smaller, more manageable subtasks. A common approach to HL, is to provide the agent with a number of high-level skills that solve small parts of the overall problem. A major open question, however, is how to identify a suitable set of reusable skills. We propose a principled approach that uses human demonstrations to infer a set of subgoals based on changes in the demonstration dynamics. Using these subgoals, we decompose the learning problem into an abstract high-level representation and a set of low-level subtasks. The abstract description captures the overall problem structure, while subtasks capture desired skills. We demonstrate that we can jointly optimize over both levels of learning. We show that the resulting method significantly outperforms previous baselines on two challenging problems: the Atari 2600 game Montezuma's Revenge, and a simulated robotics problem moving the ant robot through a maze.
\end{abstract}

\section{Introduction}


 
The \emph{options} framework by \cite{Sutton99} has emerged as the standard to tackle temporal abstractions. Previous work has shown that the use of options can improve learning efficiency in both planning and on-line learning settings \cite{mann2015approximate,fruit2017exploration}. A key question, however, is how to identify a set of suitable options for a given problem. Ideally, options should encode reusable skills that can be combined into policies for solving a wide range of related problems. A number of authors have recently investigated the problem of finding a suitable set of options in high-dimensional settings \cite{vezhnevets2017feudal,bacon2017option,kulkarni2016hierarchical}.

Most option-learning approaches rely on heuristics to identify good options. This can mean solving similar problems and identifying reusable sub-policies \cite{Konidaris09b}, identifying key states in problem transition graphs \cite{mcgovern2001automatic,csimcsek2005identifying}, or manually decomposing the problem into a set of subproblems \cite{kulkarni2016hierarchical}. The issue with these methods is that they first design a set of options and then attempt to solve the problem using these options. This requires manual option design or means that they cannot guarantee that the set of options can efficiently solve the target problem. In contrast, we propose a top-down approach that first learns the structure of the problem and then defines the set of options necessary for solving it. 
We achieve this by using a small set of human demonstrations to identify an abstract representation of the problem. 

Our method results in a set of options that can efficiently solve the overall problem. The option policies themselves can be learned using basic learners that are not able to solve the complete problem. We also reduce the problem of learning a policy over options to a small discrete problem that can be efficiently solved using basic approaches such as tabular Q-learning. These improvements mean that our method reaches human-level performance with significantly fewer samples than previous state-of-the-art methods. We demonstrate such a reduction in sample complexity on two challenging problems, where current techniques fail. Namely, we show that our method significantly outperforms state-of-the-art on the Atari 2600 game of Montezuma's Revenge and on a simulated ant robot in a maze. 


\section{Background}
\subsection{Reinforcement Learning}
We assume the standard reinforcement learning setting~\cite{sutton2018introduction} described by an MDP $\M =
(\S, \A, \p, \r, \gamma)$, where $\S$ is the (possible infinite) set of states, $\A$ the set
of (discrete) actions; $\p:\S\times\A\times \S \rightarrow [0,1]$ is the
transition function that specifies the system dynamics $\p(s'|s, a)$,
denoting the probability of transitioning to state $s'$ upon taking
action $a$ in $s$; $\r : \S\times\A \rightarrow
\Real$ is the reward function, and $\gamma$  a discount factor.

A policy is a probabilistic mapping from states to actions. For a policy $\pi$, and  MDP transition probabilities $\p$, let the matrix $\p^\pi$ denote the dynamics of the induced Markov chain: $\p^\pi(s,s') = \sum_{a \in \A}\pi(s,a)\p(s'|s,a),$ and $\r^\pi$ the reward expected for each
state under $\pi$: $r^\pi(s) = \sum_{a \in \A} \pi(s,a) \r(s,a)$. The value of a policy $\pi$ denotes the expected sum of discounted rewards starting from a given state: $\V^\pi(s) = \E_{\pi}[\sum_{k=0}^{\infty} \gamma^k R_{t+k+1} \mid S_t = s]$, where we use $S_t$ and $R_t$ to denote the state visited and reward received at time $t$, respectively. The goal of a reinforcement learner is to identify a policy $\pi^*$ that maximizes the expected discounted reward starting from any state $s$: $\pi^* = \argmax_{\pi} \V^{\pi}(s), \forall s \in \S$.

\subsection{The Options Framework}

We model high-level skills using the options framework \cite{Sutton99}. An {\it option} $o$ is a temporally extended action defined by three components: $\pi^o$, the {\it option policy} that is executed to select actions when the option is invoked, $\I^o$, the {\it initiation set} consisting of the states where the option can be started, and $\beta^o (s) \rightarrow [0, 1]$, the {\it termination condition}, which returns the probability that the option will terminate upon reaching state $s$.

{\bf Subgoal Options}: We are particularly interested in a special class of options called {\it subgoal options} \cite{Precup00a}, where the distribution over termination states (referred to as the subgoal) is independent of the distribution over starting states from which they are executed. An example of a subgoal option is navigating to a goal location; regardless of the starting state, activating the option results in the same target termination condition.

{\it Abstract subgoal options} \cite{konidaris2014constructing} model the more general case where executing an option leads to a subgoal for a subset of the state variables (called the {\it mask}), leaving the rest unchanged. For example, moving a robot's arm to a target position will not change the location of the robot itself. In other words, the state vector is partitioned into two parts $s = [a, b]$, such that executing $o$ leaves the agent in $s' = [a, b']$, where $P(b')$ is independent of the distribution over starting states.

\section{Identifying Subgoals from Demonstration}
In this section we describe the approach that we use to build high-level representations which can be used to guide behavior in large environments with sparse rewards. 
There are many existing techniques that have used demonstrations to bootstrap learning~\cite{schaal1997learning,ng2000algorithms} with some success. Rather than directly following demonstrations, the aim in our framework is to identify a small set of subgoals which can be used to guide the agent intrinsically. This leads to learning more robust policies than those obtained by trying to infer policies directly from the demonstrations. In the following section we introduce our approach for identifying subgoals.

\subsection{Probabilistic Modeling of Trajectories} 
Our model first maps a continuous set of $\bm{y}_{1},\dots, \bm{y}_{T}$ observations, to a set of continuous latent states $\bm{x}_{1},\dots, \bm{x}_{T}$. Such a mapping can be achieved using a feature extraction, such as a convolutional neural network or a variational auto-encoder in case of image trajectories. We now attempt to detect discrete switches in the trajectories, by assuming the existence of a latent state $\bm{z}_{t} \in \{1,\dots, K\}$ that evolves according to some Markovian dynamics:
\begin{equation*}
\bm{z}_{t+1}| \bm{z}_{t}, \{\mathbb{P}_{k}\}_{k=1}^{K} \sim \mathbb{P}_{\bm{z}_{t}},
\end{equation*} 
where $\{\mathbb{P}_{k}\}_{k=1}^{K}$ is the Markov transition matrix, and $\mathbb{P}_{k} \in [0,1]^{K}$ is its $k^{th}$ row. We assume that the continuous latent state $\bm{x}_{t} \in \mathbb{R}^{d_{1}}$ follows conditionally linear dynamics, where the discrete state $\bm{z}_{t}$ determines the linear dynamical system used at time t: 
\begin{equation*}
\bm{x}_{t+1} = \bm{A}_{\bm{z}_{t+1}} \bm{x}_{t} + \bm{b}_{\bm{z}_{t+1}} + \bm{v}_{t}, 
\end{equation*}
with $\bm{v}_{t} \sim \mathcal{N}(\bm{0},\bm{Q}_{\bm{z}_{t+1}})$, for matrices $\bm{A}_{k}, \bm{Q}_{k}$ both $\in \mathbb{R}^{d_{1}}$ and vectors $\bm{b}_{k} \in \mathbb{R}^{d_{1}}$ for $k = 1,2,\dots, K$. Hence, to determine $\bm{y}_{t} \in \mathbb{R}^{d_{2}}$ we can assume:
\begin{equation*}
\bm{y}_{t} = f_{\bm{z}_{t}}(\bm{x}_{t}) + \bm{w}_{t}, \ \ \bm{w}_{t} \sim \mathcal{N}(\bm{0},\bm{S}_{\bm{z}_{t}}),
\end{equation*} 
for some nonlinear function $f_{\bm{z}_{t}}$, e.g., a deconvolutional network. For the linear setting, we can easily specialize the above definition of $\bm{y}_{t}$ to:
\begin{equation*}
\bm{y}_{t} = \bm{C}_{\bm{z}_{t}} \bm{x}_{t} + \bm{d}_{z_{t}} + \bm{w}_{t},
\end{equation*}
for $\bm{C}_{k} \in \mathbb{R}^{d_{2} \times d_{1}}$ and $\bm{d}_{k} \in \mathbb{R}^{d_{2}}$. As such, the system of parameters of our model comprise the discrete Markov transition matrix and the library of linear dynamical systems, as well as the parameters of the nonlinear functions which we write as:
\begin{equation*}
\bm{\theta} = \left\{\left(\mathbb{P}_{k},\bm{A}_{k},\bm{Q}_{k},\bm{b}_{k},\bm{\theta}^{(f)}_{k}\right)\right\}_{k=1}^{K}.
\end{equation*}

\subsection{State Space Factorization}
Extracting linear dynamics from trajectories may not be feasible in cases where the trajectories are provided only as a sequence of image observations, i.e.\ it is very difficult to detect switches in a sequence of images by only reasoning about their pixels. In this work, we do not focus on the problem of visual feature extraction, but rather rely on the manually designed set of features also used in \cite{kulkarni2016hierarchical}. Recent work has demonstrated that the Transition State Clustering algorithm we use is quite robust and can be combined with various pretrained feature extraction pipelines \cite{murali2016tsc}.

With extracted features, it is generally possible identify a factored problem representation. This means that for a set of high-level features, only a subset will change at any given time. For instance, robots can first move to a goal before commencing movement of their arms to pick up an object. They do not move their arms and change their position at the same time. We utilize this observation to reduce the dimensionality of subgoal identification, by splitting features into subsets and detecting subgoals separately within each of these subsets. This has the additional advantage that the learned subgoals now become abstract subgoals in the context of the original problem. This means that we can learn a policy to achieve a subgoal without affecting the status of unrelated subgoals. In the next section, we exploit this fact to decompose the learning problem.

More formally, we use the feature trajectories to identify the set of feature \emph{factors} $F$, where each factor $f \in F$ represents a set of features that covary. For example, in our Montezuma's Revenge experiments we extract the agent's $x$ and $y$ position, and whether the key has been obtained (these are the features which are most relevant for determining the agent's reward), which leads to two identified factors: the agent's $(x,y)$ position, and the status of the key. In the next section we introduce the method we use to identify subgoals. 

\subsection{Transition State Clustering}

The problem discussed in Section 3.1 is one of detecting switches in linear dynamical systems. In this paper, we instantiate a solver by relying on the Transition State Clustering (TSC) algorithm \cite{krishnan2017transition}. This algorithm segments a set of trajectories by first identifying the switch-points (and corresponding switch-times) corresponding to the changes in linear dynamics within each trajectory, and then clustering the switch-points from the entire set of trajectories.

As mentioned in the previous section, we identify the set of abstract subgoals represented in the demonstrations by running TSC separately for each factor. We slightly modify TSC by adding one step: for each switch-time discovered by the first step of TSC, add the switch-time to the other factors in the same trajectory (with corresponding switch-points). For example, in Montezuma's revenge we would add transitions corresponding to the agent's last position before obtaining the key. This modification ensures that the abstract subgoal property holds for the set of discovered subgoals, i.e.\ that it is possible to achieve the subgoals without affecting the status of unrelated subgoals.

\section{Problem Decomposition}
Our goal in this section is to decompose the MDP into a set of simpler MDPs. To achieve this, we first define a top-level \emph{abstract MDP} \cite{marthi2007automatic}. This abstract description captures the overall problem structure and uses an abstracted state space that considers subgoals rather than atomic states. The actions of the abstract MDP are options that achieve one of the subgoals. Since we have constrained our subgoals to be \emph{abstract subgoals}, we can assume that it is possible to learn options to achieve each subgoal, without changing state variables that do not relate to the subgoal (i.e. are not in the same factor).  Based on the abstract MDP description,  we then define a set of low-level base MDPs. Each of these MDPs describes the learning problem for one of the option policies used by the abstract MDP. We can then learn a solution in the abstract MDP to find the sequence of options necessary to solve the problem, and solve each of the low-level MDPs to find the option policies.

\subsection{Abstract MDP}
To define the abstract MDP we start from the set of subgoals from Section 3.3: $H_f = \left \{h^f_1, \ldots, h^f_{|H_f|} \right \}, \forall f \in F$ from demonstration trajectories. The abstract MDP $\mu$ describes the problem using an abstracted state space and temporally extended actions. We use the subgoals to define the \emph{abstract MDP $\mu = (\S^\mu, \A^\mu, \r^{\mu},\p^\mu,\gamma)$} as follows:
\begin{itemize}
\item The state space of the MDP is the product space of the abstract subgoals for each of the factors: $S^{\mu} = H_{f_1} \times ... \times H_{f_n}$.
\item The action space consists of a set of abstract subgoal options. One option for each subgoal identified; $A^{\mu} = \{ o_{h_i} \mid h_i \in H_f, f \in F \}$. Each option $o_h$ is defined as the tuple
$o_h = (\beta^{o_h}, \pi^{o_h}, \I^{o_h})$.  The termination function $\beta^{o_h}$ is $\indic{h}$, the indicator function for the subgoal $h$. The option policy $\pi^{o_h}$ is learned as described below. The initiation set $\I^{o_h}$ is the entire abstract MDP state space $S^\mu$ except $\indic{h}$ (we cannot initiate the option to achieve subgoal $h$ in states where that subgoal has already been achieved).
\item The reward $\r^\mu (s^\mu, o, {s^\mu}')$ that the meta-controller observes from executing an option $o$ at starting time $t$ in abstract state $s^\mu$ and terminating after $T$ steps in abstract state ${s^\mu}'$, is the expected cumulative discounted environmental reward observed during the execution of $o$ in the original MDP: $\r^\mu(s^\mu,o,{s^\mu}') = \E_{\pi^o} [ \sum_{k=0}^{T} \gamma^k \R_{t+k} ]$.
\end{itemize}

\subsection{Base MDPs}
For each subgoal option $o_h$ defined in the abstract MDP, we consider a distinct base MDP $ M^{o_h} = (\S, \A, \r^{o_h}, \p, \gamma) $. These MDPs use the same state space $\S$, action space $\A$ and transition function $\p$ as the original MDP, but replace the reward function $\r$ with a new intrinsic reward $\r^{o_h}$ that defines the option learning task.

A simple reward function to learn the subgoal options is the indicator function for the subgoal $h$. However, this would lead to a very sparse reward signal, so we use a more informative potential based \cite{ng1999policy} reward signal for each option, $r^{o_h}(s,a,s')$, which provides a reward based on the reduction in distance to the goal state plus a large bonus when the goal state is achieved. This goal-based reward in combination with a discount $\gamma \in (0,1)$ is sufficient to define a learning problem where the optimal policy is to reach the subgoal as quickly as possible.

Other literature has proposed using some combination of intrinsic and environmental reward when training controllers for options \cite{kulkarni2016hierarchical}. We do not do this because one of the primary benefits of learning options is that they can be reused when solving different tasks within the same environment, which could involve a different environmental reward function.

\subsection{Integrated architecture}
Overall, our goal is to maximize expected reward over both the meta-policy $\pi^\mu$ and the set option policies $\Omega = \{\pi^{o_h}, \forall o_h \in A^\mu\}$. Let $\xi(o_h,s,t)$ be the event that option $o_h$ is initiated in state $s$ at time $t$. Then the complete optimization problem is:
$$\underset{\pi^\mu}{\max} ~ \underset{\Omega}{\max} ~ \E_{\pi^\mu, \Omega} \Bigg[ \sum_{t=0}^T  \gamma^t R_t \mid \xi(o_h,s,t) \Bigg]$$
As mentioned in the previous section, optimizing the intrinsic reward leads to option policies which reach their corresponding subgoals as quickly as possible. Therefore, if the environmental reward signal only occurs when a subgoal is reached, then the joint optimization of the meta and low-level controllers will lead to optimal behavior in the underlying MDP.

While it is possible so solve each of the MDPs defined above in isolation and then solve the abstract MDP, this would result in a very inefficient learning method. In practice, we consider a setup where both the abstract MDP and base MDPs are solved jointly. We use the call-and-return option model; when an option is invoked, the agent chooses low-level actions according to the corresponding option policy until the option terminates, and another option can be chosen. In our setting, an option terminates only when its abstract subgoal is reached (determined by a distance threshold), the option times out (executes a certain number of steps without achieving its subgoal), or the episode ends (as determined by the agent's environment).

Training proceeds as follows: the meta-controller chooses an option $o_h$ to execute from its current abstract state $s^\mu$. Actions are then chosen using the policy of option $o_h$, which is updated with observed transitions $(s,a,s',r^{o_h})$. When a terminal condition is reached, we update the meta-controller policy with transition $(s^\mu,o_h,{s^\mu}', r^\mu(s^\mu,o_h,{s^\mu}'))$, where ${s^\mu}'$ is the terminal abstract state. See Algorithm 1 for a complete description of our approach. We note that our approach is actually agnostic to the learning algorithm used to solve the base MDPs. In the following section we show that the decomposition allows us to use low-level controllers that are unable to solve to original full MDP $\M$.

\begin{algorithm}
    \SetKwInOut{Input}{Input}
    \SetKwInOut{Output}{Output}

    \Input{Abstract state space and set of untrained options}
    \While{Training is not over}
      {
      
      choose option $o_h$ according to $\pi^\mu$\;
      $t=0$, $r^\mu = 0$, $s^\mu= $ starting abstract state\;
      \While{Option or episode not terminal}
      	{
        choose action $a$ according to $\pi^{o_h}$\;
        observe transition $(s,a,s',r^{o_h}, r)$\;
        update $\pi^{o_h}$ with $(s,a,s',r^{o_h})$\;
        $r^\mu = r^\mu + \gamma^t r$ \;
        $t = t + 1$\;
        }
        ${s^\mu}'=$ terminal abstract state\;
        update $\pi^\mu$ with $(s^\mu,o_h,{s^\mu}', r^\mu)$\;
        \If{episode terminal}
        {
        reset environment\;
        }
      }
    \caption{Jointly training meta-controller and options}
\end{algorithm}

\section{Experiments}
We evaluate our approach in $2$ challenging settings: an Atari 2600 game and a simulated robotics problem.\footnote{Videos of the results of training are available at: \url{https://sites.google.com/view/learningrepresentations}} In order to demonstrate the flexibility of the approach, we vary the methods used to solve the low-level skill learning  problems; we use DQN in the Atari domain and PPO in the robotics domain. We show that our problem decomposition allows our reinforcement learner to efficiently solve the problem, even though they the basic algorithms are not able to solve the complete problem. Moreover, we show that the approach significantly outperforms state-of-the-art baselines in terms of sample complexity and final performance.

\subsection{Montezuma's Revenge}

\begin{figure}[h]
    \centering
    \includegraphics[width=0.9\linewidth]{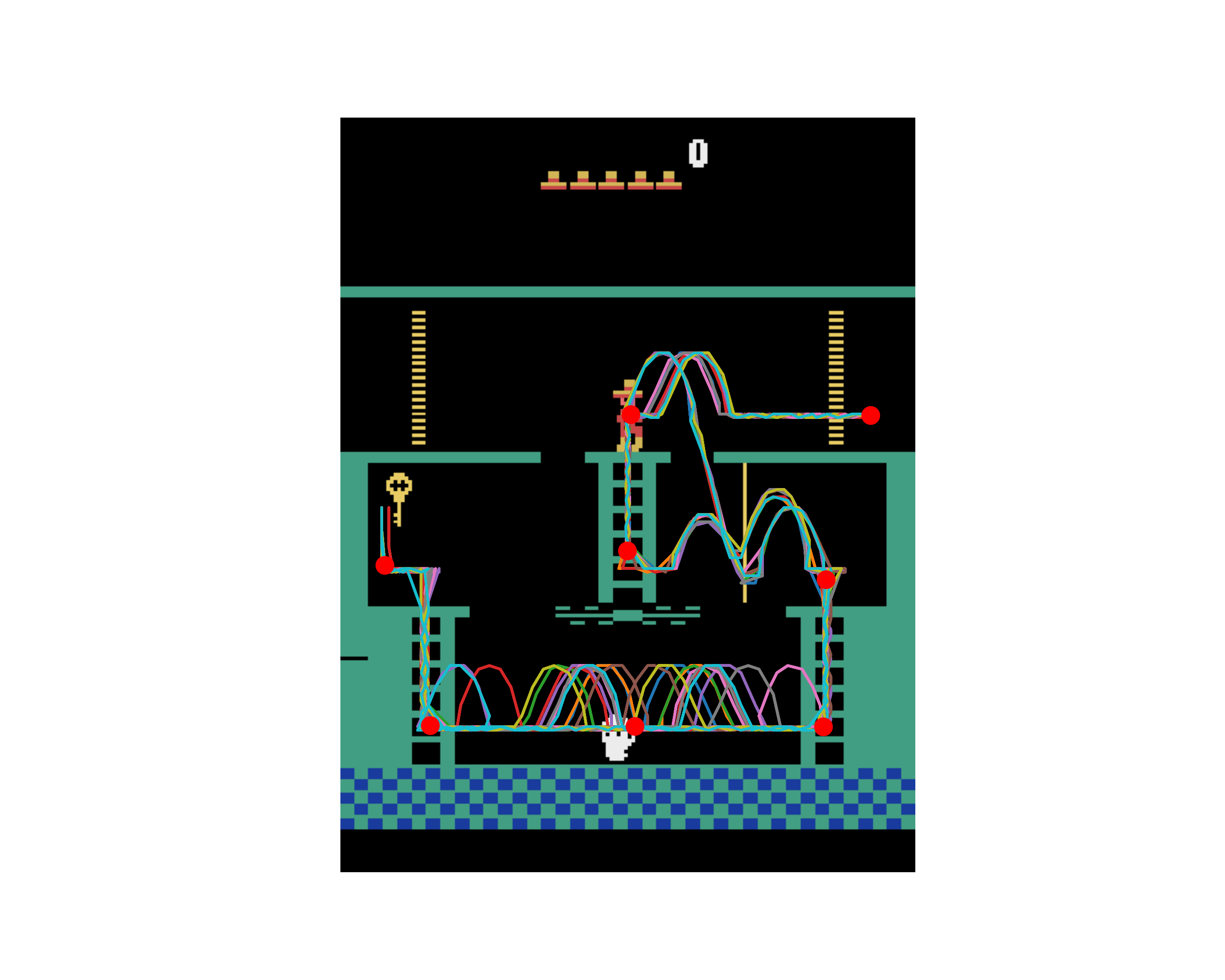}
    \caption{The first room of Montezuma's Revenge with demonstration trajectories and discovered position abstract subgoals (red dots). Not shown is the abstract subgoal for obtaining the key.}
    \label{fig:mont}
\end{figure}

We first demonstrate our algorithm on the first room within the Atari 2600 game Montezuma's Revenge (the environment is reset whenever the agent opens one of the doors). We obtained the 10 demonstration trajectories shown in Figure \ref{fig:mont}. As mentioned in Section 3.2, we use a manually defined feature extractor to determine the position of the agent and whether the key was present in the frame. This resulted in the identification of two factors: one being the $(x,y)$ position of the agent and one being the status of the key. We run the TSC algorithm using 10 copies of each of these trajectories to obtain the position abstract subgoals shown in Figure \ref{fig:mont}, plus a key-status abstract subgoal which is not shown.

As discussed in the previous section, we used these discovered abstract subgoals to define low-level controllers and then jointly learn policies for these low-level controllers along with the meta-controller policy. In practice, these subgoals result in the low-level controllers learning to move the agent to specific locations and to pickup the key. 

We use independent Deep Q-Networks (DQNs) \cite{mnih2015human} to learn policies for achieving each abstract subgoal\footnote{It seems desirable to share features between the networks in some way, however we found the performance of independent networks to be vastly superior. Other work which uses shared features also experienced difficulty with training.} and tabular Q-learning to train the meta-controller (a key benefit of our approach is that it leads to very simple and efficient learning at the meta level). Each option has its own replay buffer; the agent allocates experiences to the replay buffer of the option that is currently executing. Additionally, if the agent enters a terminal state for an option other than the one it is currently executing, the cumulative experience since the previous terminal state is added to that option's replay buffer.  

\begin{figure}[h]
          \centering
          \includegraphics[width=\linewidth]{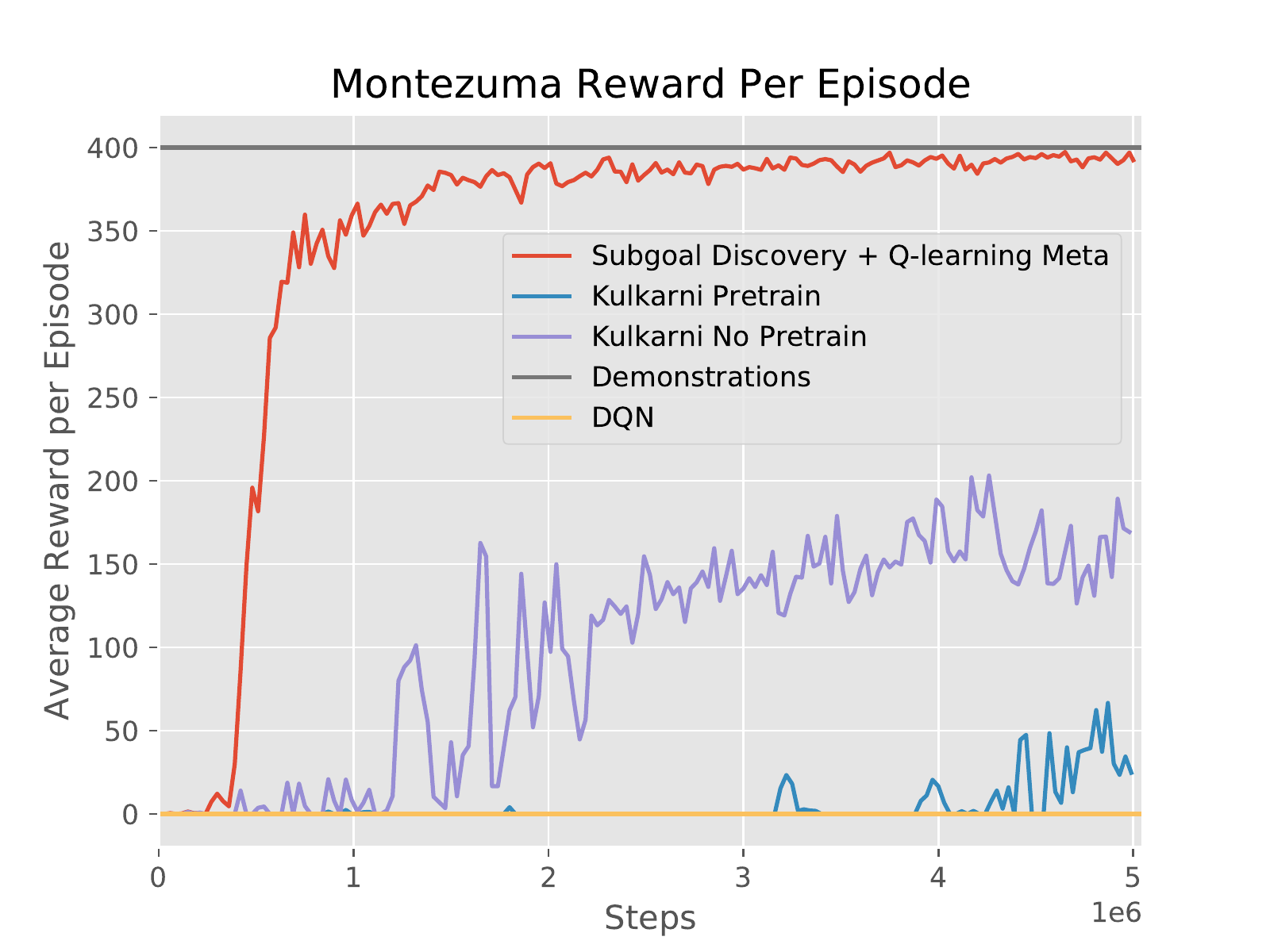}         
          \caption{Simulation results for the first room of Montezuma's Revenge. The maximum reward that can be obtained from a single episode is 400; 100 for obtaining the key and 300 for opening a door. We compare our method against the performance of the demonstrations, basic DQN, and the Kulkarni approach with and without the pretraining phase. Our approach significantly outperforms the baselines, achieving human-level performance in 2 million training steps.}
        \label{fig:keydoorperepisode}
        \end{figure}

In Figure \ref{fig:keydoorperepisode} we compare our method against basic DQN, and the hierarchical Deep Q-learning approach proposed by Kulkarni et al.\ \cite{kulkarni2016hierarchical}. Kulkarni et al.\ used a pretraining step that trains the low-level controllers for 2.5 million steps before moving to the full problem. We show results for their algorithm both with and without this phase.\footnote{\url{https://github.com/mrkulk/hierarchical-deep-RL}} Our approach significantly outperforms these baselines and attains human-level performance in 2 million training steps.

\subsection{Ant Robot in a Maze}

\begin{figure}[h]
    \centering
    \includegraphics[width=.8\linewidth]{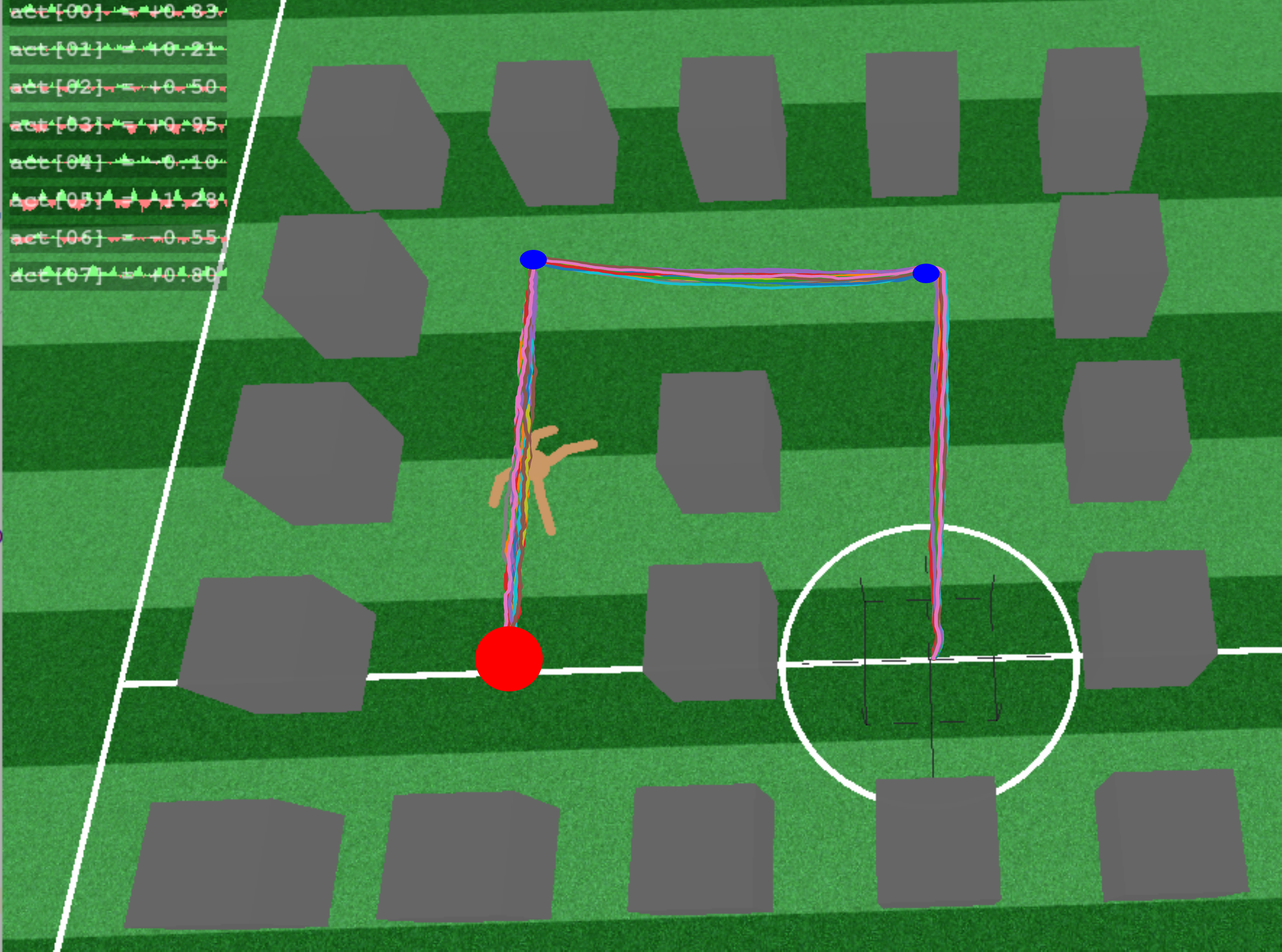}
    \caption{The ant maze simulation environment with demonstration trajectories and discovered subgoals (blue dots). The robot's starting location is the bottom right of the maze and its goal is to navigate to the bottom left of the maze to the goal region (red circle).}
    \label{fig:antmaze}
\end{figure}

The second environment we consider is an ant robot in a maze, shown in Figure \ref{fig:antmaze}. The robot is simulated using the Roboschool\footnote{\url{https://github.com/openai/roboschool}} physics simulator, and has 8 and 28 dimensional continuous state and action spaces, respectively. The goal of the robot is to navigate to the end of the maze shown in Figure \ref{fig:antmaze}, which returns an environmental reward of 1. The minimum number of actions required to reach the goal is approximately 1200. While recently proposed algorithms such as Proximal Policy Optimization (PPO) \cite{schulman2017proximal} have made great progress in basic locomotion tasks, empirical evaluations show that they fail in more complex navigation tasks \cite{duan2016benchmarking}. A fundamental issue is that the algorithms are trained using a potential-based reward signal and a relative state-space, which have difficulty capturing the hierarchical nature of navigating a maze. Our method can utilize these algorithms to train the option policies, while still managing the hierarchical aspect of the task with the meta-controller.  

We generated 15 demonstration trajectories (also shown in Figure \ref{fig:antmaze}) by training a sequence of low-level controllers with predefined subgoals. From the trajectories we identified one factor which was the $(x,y)$ position of the robot. We ran TSC to obtain two additional subgoals to the original goal, shown in Figure \ref{fig:antmaze}. The meta-controller was trained using tabular Q-learning and the low-level controllers were trained using PPO, with independent policy networks for each option. 

In addition to demonstrating learning from scratch, we also show the effect of injecting prior knowledge into the meta-learning process. We consider 2 cases. First, we assume that the high-level structure of the problem is completely known (i.e. we know the sequence of subgoals that need to be achieved). In this case, the meta-controller is fixed and there is no learning at the meta-level. This setting similar to the problem considered in the Policy Sketches approach \cite{andreas2016modular}. In the second case, we initialize the meta-learner using the demonstration trajectories. We use probabilistic policy-reuse \cite{fernandez2006probabilistic} for the meta-controller, which works by copying the high-level action that was observed in the demonstrations at each subgoal with some probability. 

\begin{figure}[h]
    \centering
    \includegraphics[width=\linewidth]{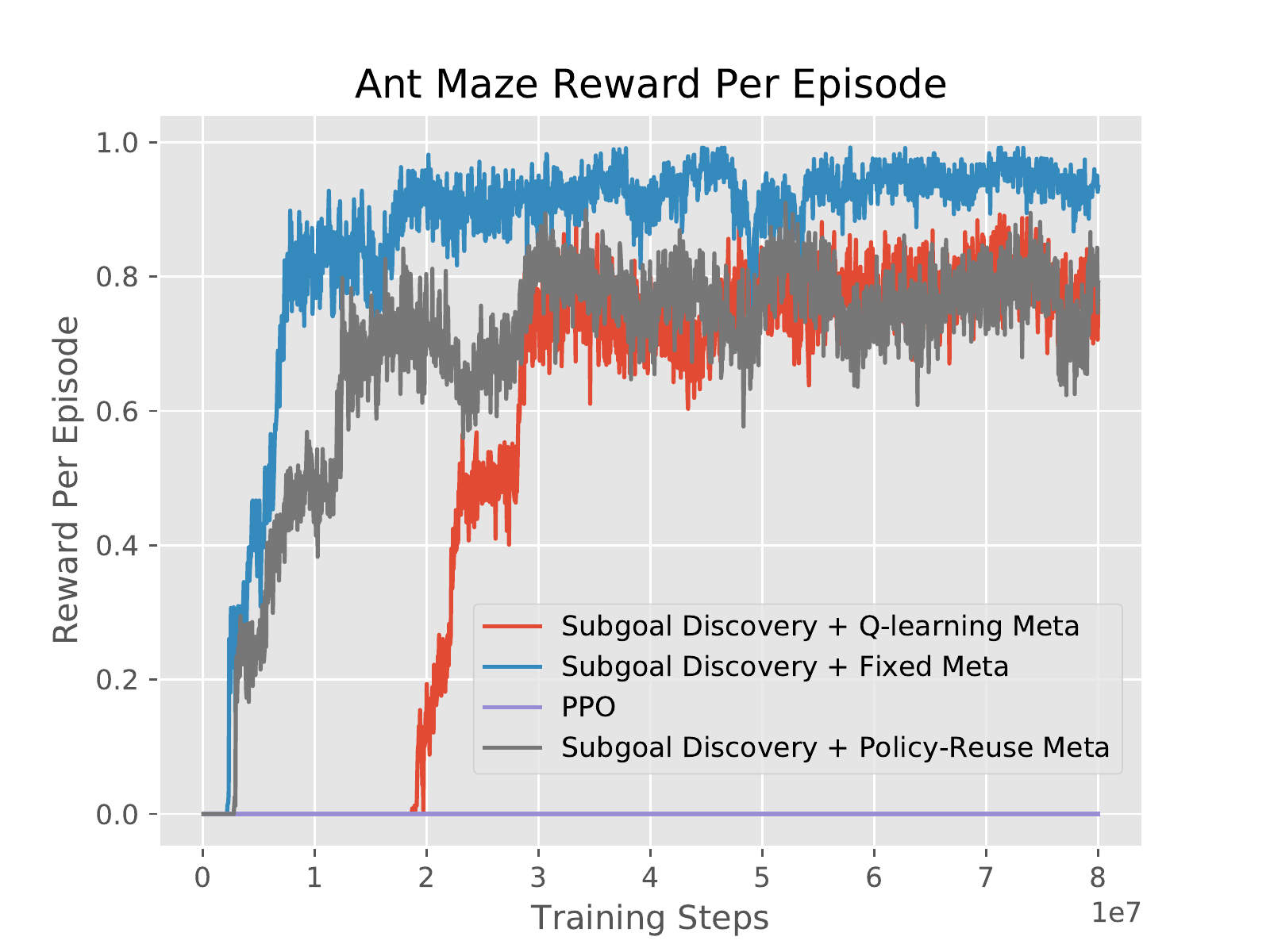}
    \caption{Simulation results for the ant maze domain. We compared our subgoal discovery method combined with a Q-learning meta-controller, a probabilistic policy-reuse meta-controller, and a fixed meta-controller, against PPO. The fixed meta-controller learns the fastest because it only needs to learn the low-level policies. Probabilistic policy-reuse interpolates between Q-learning and the fixed meta-controller. The Q-learning and policy-reuse meta-controllers have worse maximum performance than the fixed meta-controller because they use $\epsilon = 0.1$ at the meta-level.}
    \label{fig:antep}
\end{figure}

We compare the resulting four approaches: basic PPO, our method, our method with probabilistic policy-reuse, and a fixed meta-controller with the discovered subgoals. Each algorithm had access to the original goal, and all algorithms besides PPO additionally used the same discovered subgoals. Figure \ref{fig:antep} shows the results of our experiments. Our method was able to solve the maze, even though basic PPO was never able to. Additionally, adding probabilistic policy-reuse to our method significantly increases the learning speed.

\noindent {\bf Hyperparameter Settings:} For both environments, we used $\gamma = 0.99$. For the meta-controllers of both environments, we used random action probability $\epsilon = 0.1$, and learning rate 0.2. For probabilistic policy reuse, we linearly decayed the probability of copying the demonstrations from 0.9 to 0 over 5 million steps. For the Montezuma low level DQNs we used the OpenAI baselines\footnote{\url{https://github.com/openai/baselines}} implementation with default settings except for learning rate 0.001, target network update frequency 10,000 steps, size of each replay buffer 200,000 steps, and a linearly decayed $\epsilon$ from 1 to 0.1 over 75,000 steps independently for each controller. For the ant robot low-level controllers, we used the OpenAI baselines implementation of PPO with the default settings except for independent learning rates for each controller that were linearly decayed from 1 to 0.1 over 2 million steps. We determined that a subgoal was reached if the agent's distance was less than 6 and 0.5 for Montezuma and ant, respectively. In both domains, the low-level controllers time out after 1000 steps.


\section{Discussion and related work}
Hierarchical approaches have a long history within the reinforcement learning literature. Many approaches have been proposed to decompose difficult learning problems into a set of smaller subproblems, see e.g. \cite{wiering1997hq,Precup00a,dietterich2000hierarchical}. In this section we mainly focus on the options framework as proposed in \cite{Precup00a}, in particular we look at approaches that combine high dimensional inputs and hierarchies.

Several papers have recently tried to combine deep reinforcement learning and hierarchical learning. These works typically focus on integrating temporally extended actions with deep networks architectures. Kulkarni et al. \cite{kulkarni2016hierarchical} extend the basic DQN algorithm \cite{mnih2015human} with a set of heuristic, predefined options. They show that the resulting method is able to solve the first room of Montezuma's Revenge. A key issue with this approach is that it does not provide a clear way to determine a suitable set of options. Moreover, it requires a separate pretraining phase to first learn the option policies before being able to solve the main problem. In our Montezuma experiments, we provide the performance of this method as a baseline.

Other methods attempt to learn both options and the policy over options at the same time.  Bacon et al. \cite{bacon2017option} introduce the end-to-end Option-Critic architecture. They formulate a policy gradient method capable of learning the option policies, option terminations and policy-over-options. 
A key issue with these and other option-learning approaches is that they tend to degenerate to trivial solutions that either use a single option policy to solve the entire problem, or use only primitive actions (i.e. they either never switch options or switch options on every step). While these methods are able to the learning problem, they fail to identify meaningful subgoals that lead to useful abstractions and reusable options. FeUdal Networks \cite{vezhnevets2017feudal} attempt to solve this issue by decoupling the option learning from the higher-level learning.  They propose a 2-level hierarchical network approach where the Manager network sets goals and intrinsic rewards for the lower level Worker network. The authors demonstrate that the approach is able to learn meaningful subgoals in the first room of Montezuma's Revenge.

Superficially, all the approaches above consider a high-level/low-level architecture, similar to our meta-level controller choosing sub-goals for the base-level controllers. Our approach has a number of key advantages, however. First, we operate in top-down fashion by identifying meaningful subgoals from human demonstrations. This means that we do not rely on predefined options, but rather can identify the set of options necessary to solve the problem. Additionally, we can ensure that these are meaningful options that do not degenerate into trivial solutions. Second, unlike the end-to-end deep hierarchical approaches, our meta-level controller learns on an abstract level that decouples the low-level representations (e.g. pixels) from the high-level problem structure. This has the advantage that the meta-controller does not need to work with high-dimensional inputs, but can operate on a simplified, symbolic level. This allows a more efficient solution to the overall learning problem. Moreover, it means the high-level controller  becomes agnostic to the low-level algorithms used to solve the base MDPs. Finally, as shown in the Roboschool experiments, we can combine the low-level skill learning with various amounts of prior knowledge about the overall problem structure. While not required for successful learning, any knowledge injected into the process can significantly speed up learning.

\section{Conclusion}
In this paper, we proposed a method to use  demonstrations in order to decompose complex learning problems into a set of simpler subtasks. We achieve this by identifying a set of subgoals and relations between them, and then learn a set of options to solve each of the identified subproblems.  We empirically demonstrate that our method significantly increases the learning speed over previous approaches on two challenging problems: the Atari 2600 game Montezuma's Revenge and a simulated robotics problem moving the ant robot through a maze.

\bibliographystyle{named}
\bibliography{sym-prob}{}

\end{document}